\documentclass[sigconf]{acmart}
\usepackage{amsmath,epsfig,graphicx, booktabs}

\usepackage{amssymb}
\usepackage{paralist}
\usepackage[ruled,linesnumbered]{algorithm2e}

\AtBeginDocument{%
  \providecommand\BibTeX{{%
    \normalfont B\kern-0.5em{\scshape i\kern-0.25em b}\kern-0.8em\TeX}}}

\setcopyright{acmcopyright}
\copyrightyear{2021}
\acmYear{2021}
\acmDOI{10.1145/1122445.1122456}

\acmConference[Chengdu '21]{Chengdu '21: ACM International Conference on Multimedia}{October 20--24, 2021}{Chengdu, China}
\acmBooktitle{Chengdu '21: ACM International Conference on Multimedia,
  October 20--24, 2021, Chengdu, China}
\acmISBN{978-1-4503-XXXX-X/18/06}

\acmSubmissionID{407}

\begin{document}

\title{Context-aware visual tracking with joint meta-updating}


\author{Qiuhong Shen}
\affiliation{%
  \institution{Harbin Institute of Technology, Shenzhen}
  \city{Shenzhen}
  \country{China}}
\email{shenqiuhong0905@gmail.com}

\author{Xin Li}
\affiliation{
  \institution{Peng Cheng Laboratory}
  \city{Shenzhen}
  \country{China}
}
\email{xinlihitsz@gmail.com}

\author{Fanyang Meng}
\affiliation{
  \institution{Peng Cheng Laboratory}
  \city{Shenzhen}
  \country{China}
}
\email{mengfy@pcl.ac.com}

\author{Yongsheng Liang}
\affiliation{
  \institution{Harbin Institute of Technology, Shenzhen}
  \city{Shenzhen}
  \country{China}
}
\email{liangys@hit.edu.cn}







\begin{abstract}
Visual object tracking acts as a pivotal component in various emerging video applications.
Despite the numerous developments in visual tracking, existing deep trackers are still likely to fail when tracking against objects with dramatic variation.
These deep trackers usually do not perform online update or update single sub-branch of the tracking model, for which they cannot adapt to the appearance variation of objects.
Efficient updating methods are therefore crucial for tracking while previous meta-updater optimizes trackers directly over parameter space, which is prone to over-fit even collapse on longer sequences.
To address these issues, we propose a context-aware tracking model to optimize the tracker over the representation space, which jointly meta-update both branches by exploiting information along the whole sequence, such that it can avoid the over-fitting problem.  
First, we note that the embedded features of the localization branch and the box-estimation branch, focusing on the local and global information of the target, are effective complements to each other.
Based on this insight, we devise a context-aggregation module to fuse information in historical frames, followed by a context-aware module to learn affinity vectors for both branches of the tracker.
Besides, we develop a dedicated meta-learning scheme, on account of fast and stable updating with limited training samples.
The proposed tracking method achieves an EAO score of 0.514 on VOT2018 with the speed of 40FPS, demonstrating its capability of improving the accuracy and robustness of the underlying tracker with little speed drop.

%
\end{abstract}


%


\maketitle

\section{Introduction}
\label{sec:intro}
Visual object tracking has been an important part of various video applications, such as autonomous driving~\cite{ad_survey}, action recognition~\cite{ac_survey}, and video analysis~\cite{va_survey}.
Given a bounding box representing the target object in the first frame, the goal of visual tracking is to estimate the target state (location and scale) in subsequent frames. 
In recent years, deep trackers~\cite{DiMP,SiamRPN++,DMSTIT, SiamR-CNN, ROAM} aided with the powerful representation ability of CNN achieve significant progress in both terms of robustness and accuracy. 
However, how to effectively adapt a deep tracking model to target changes overtime is still challenging due to the limited online training samples and high-computational loads.

\begin{figure}[t]
    \centering
    \includegraphics[width=0.9\linewidth]{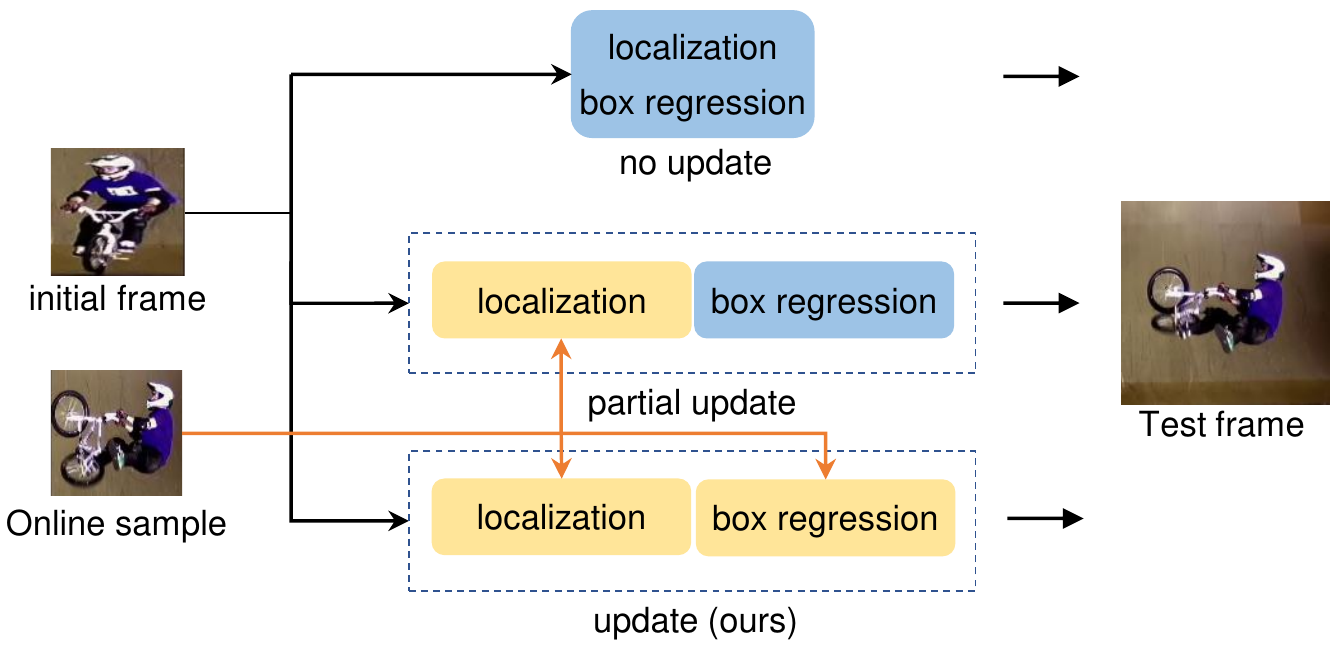}
    \caption{\textbf{Motivation of our proposed tracker.} Our approach updates tasks of localization and box-estimation jointly on the representation space.}
    \label{fig:motivation}
\end{figure}

Existing mainstream tracking methods can be roughly divided into offline-trained and online-optimized models based on whether they only learned a metric-space in offline stage or perform online model optimization(weight update).
\begin{figure}[b]
\centering
\includegraphics[width=0.95\linewidth]{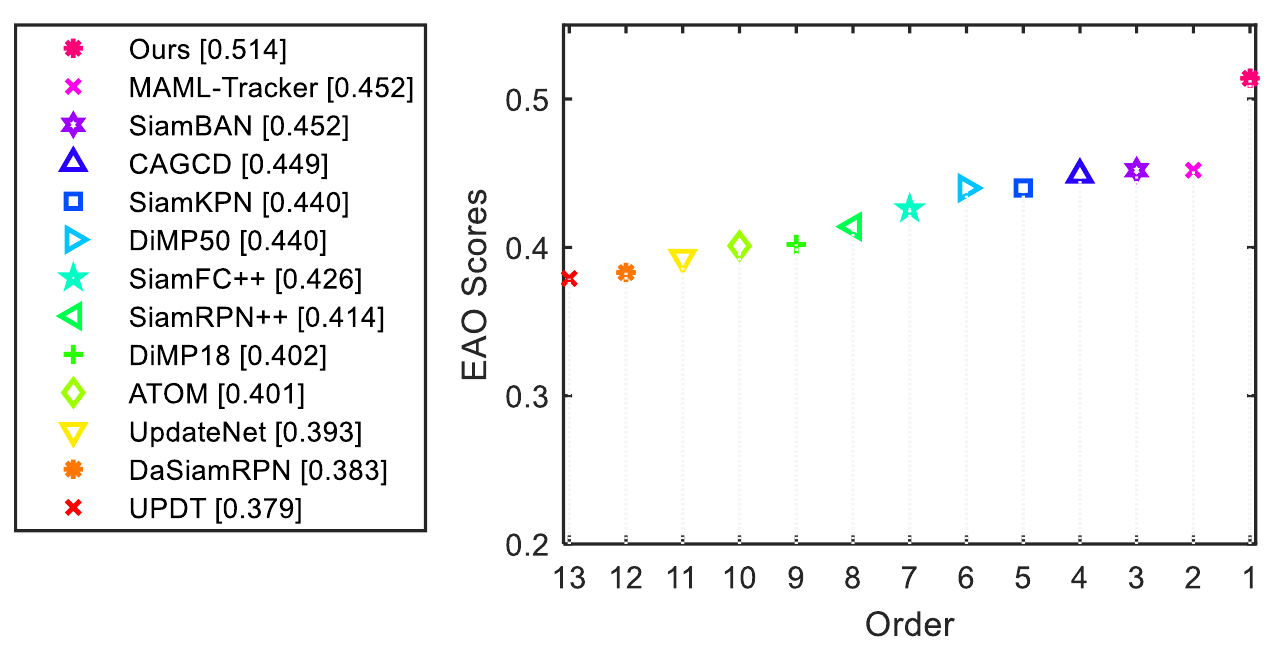}
\caption{\textbf{Comparison with state-of-the-art on VOT2018~\cite{VOT2018}.} Better viewed in color. Our method achieves an EAO of 0.514, which significantly outperforms baseline(DiMP~\cite{DiMP}) with a relative gain of 16.8\%. }
\label{fig:eao_subplot}
\end{figure}
The former, like SiamRPN model~\cite{SiamRPN}, develops a representative offline tracking framework that performs localization and box-estimation simultaneously using the branches of classification and bounding box regression.
Benefited from the high efficiency structure of Siamese tracker, numerous methods~\cite{DaSiamRPN,SiamRPN++,C-RPN} built under the SiamRPN framework achieve fast and accurate performance.
However, an explicit template is required in this category of trackers, which hinders effective online method to exploit variation in historical frames, resulting in the poor capability to distinguish the target from background distractors that often undergoing dramatic changes.
In contrast, another kind of deep trackers~\cite{ATOM,DiMP,PrDiMP} first localize the target with an online learned classifier and then estimate the scale with a regression module(IoU prediction).
Equipped with online learning, these trackers adaptively learn the discriminative feature between the target and the background such that they can distinguish the target from the background effectively.
Despite the robust performance, this kind of trackers only update the classification branch for rough localization but fix the regression branch, which brings about severe drifting error for targets with strong variation.

Based on the above analysis, existing deep trackers do not adapt to online targets effectively in terms of the following aspects.

First, the offline-trained methods only use the initial target sample as the template and keep it fixed during tracking. 
Since the target appearance is changed over time, it is less effective to locate the target only based on the information from the initial target sample.
Second, although the online-optimized methods perform online learning, they only update part of the sub-tasks, which leads to inaccurate target box predictions when the target appearance changes a lot.

To address the above-mentioned issues, we propose a context-aware adaptive tracking framework that adapts the tracking model to target changes jointly and effectively.

We note that the box-estimation branch focuses more on global target features, while the localization branch focuses more on the local discriminative features of the target.
This observation indicates that the learned information of the localization branch can be used to guide the adaption of the box-estimation branch and the adapted features of the localization branch help the localization branch to find the center of the target conversely.
To bridge the relationship between these two branches and update them jointly, we develop a context-aware aggregation model with an affinity vector for adjusting both branches. 
We show that the affinity vector can not only advance the localization ability but also improve the target estimation accuracy. 
In addition, we provide a gradient-based meta-learning method to update the context-aware module, which ensures an effective online adaption with limited training samples.
We evaluate the proposed algorithm on four diverse datasets including OTB100~\cite{OTB2015}, VOT2018~\cite{VOT2018}, NFS~\cite{NFS}, LaSOT~\cite{LaSOT} and TrackingNet.
The favourable performance against the state-of-the-art methods demonstrates the effectiveness of the proposed algorithm.

In this paper, we make the following contributions:

\begin{compactitem}
  \item We propose a context-aware module to update both the localization and box regression models jointly for visual tracking, which enables online adaption and interaction between two sub-tasks in visual tracking, leading to robust and accurate tracking performance.
  \item We develop a meta-learning strategy for ensuring a fast and efficient online adaption of the context-aware module with limited training samples.
  \item We conduct extensive experiments on VOT2018, OTB100, NFS, LaSOT and TrackingNet to demonstrate the effectiveness of our approach with favourable performance against the state-of-the-art methods.
\end{compactitem}

\section{Related work}
In this section, we review the closely related studies including deep trackers and online adaption models.

\begin{figure}[t]
    \centering
    \includegraphics[width=1\linewidth]{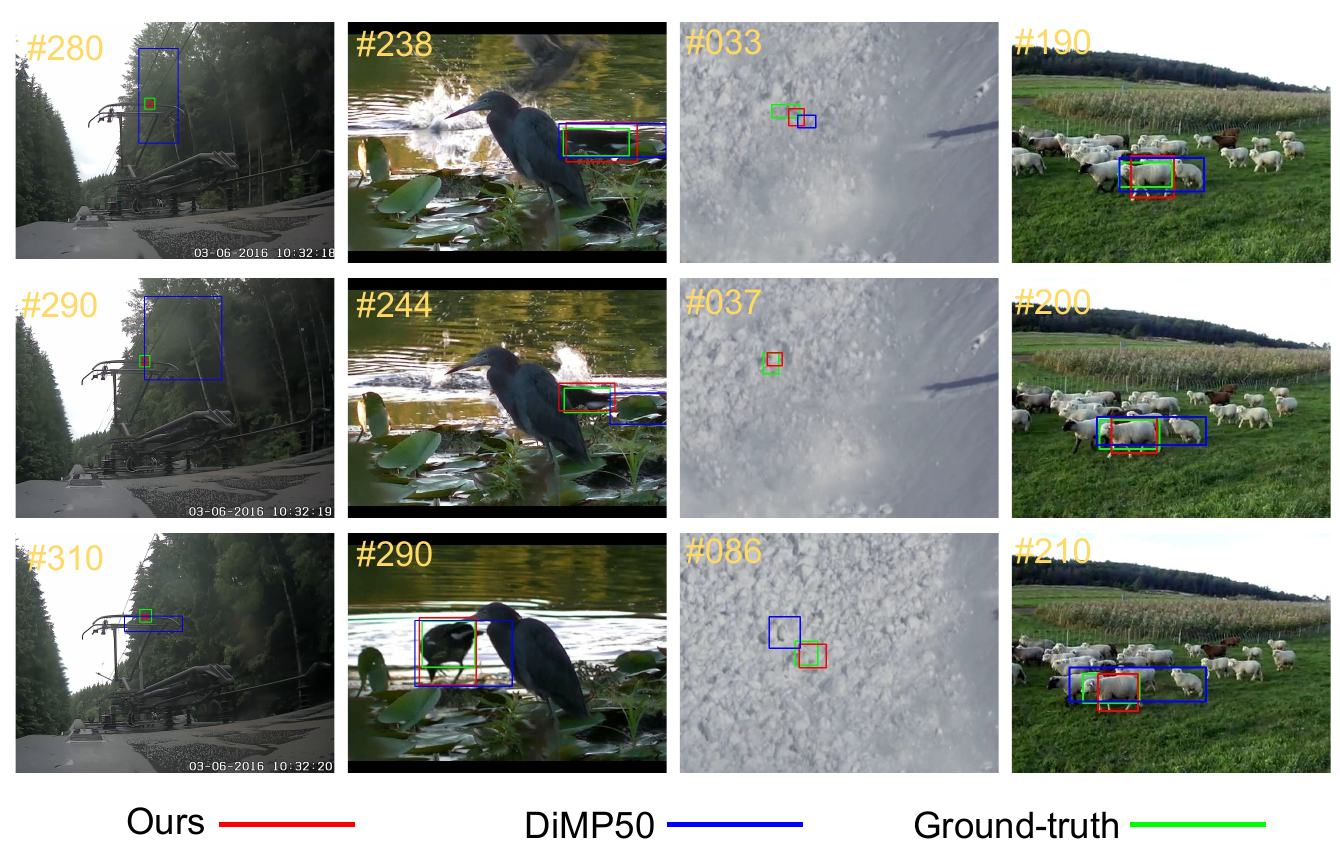}
    \caption{\textbf{Qualitative evaluation of CAJMU and DiMP50 on several challenging sequences.} Our joint meta-updater shows strong capability of discriminating distractor in the background and achieves more accurate bounding box regression, alleviating the drifting error in tracking task.}
    \label{fig:vis_tracker}
\end{figure}

\vspace{2mm}
\noindent\textbf{Deep trackers.}
With the great success of deep learning in various computer vision tasks, numerous deep learning based trackers have been developed. 
We divide them into two categories: offline-trained trackers and online-optimized trackers, depending on online learning(weight updating) is applied or not. 
Siamese network based trackers are the typical category of offline-trained tracker. SiamFC~\cite{SiamFC} is the pioneering work, which adopted a fully-convolutional siamese network for training a metric space for offline tracking in end-to-end manner. During tracking, a search patch centered at the previous positon of the target is extracted to correlation with the template feature extracted from first frame to yield a similarity map. The rough location of the target is localized find the maximum position in this similarity map, the box-estimation task is performed using a multi-scale search, which limit the performance in this model.
SiamRPN~\cite{SiamRPN} solved this drawback by borrowing RPN module from object detection task. To achieve more accurate target box predictions, the feature extractor is successfully replaced by ResNet~\cite{RES50} in SiamRPN++~\cite{SiamRPN++} via spatial aware sampling strategy to maintain translation invariance.
And advanced component borrowed from detection models~\cite{SiamRPN,C-RPN, TACT, RPT} and deeper backbones~\cite{SiamRPN++, SiamDW} are exploited under the Siamese framework.
The correlation operator is evolved in these trackers to better fuse the feature from the template and search region. Recently, graph attention~\cite{graph_track, SGT} and transformer~\cite{TransT} based fusion method are introduced to substitute original simple correlation method to further boost the tracking performance.
While the online-optimized tracking method is dominated by DCF-based trackers~\cite{ECO, CCOT, dcfnet} in the early stage. ECO~\cite{ECO} is the typical DCF-based tracker using deep features, which introduce a factorized convolution operator to reduce the number of paramters in DCF model and an efficient model update strategy to significantly improve the tracking speed of this category of trackers.
Recently, ATOM~\cite{ATOM} devised a box-estimation head to predict the overlap between the target box and candidate boxes, which achieved accurate bounding box estimation by integrating target-specific information.
Besides, these online-optimized trackers always perform classification in an online ridge regression manner to yield a rough localization of the target, then processed by a bounding box estimation module. 
To enhance the robustness to target changes, this kind of methods usually exploit a predictor learned online~\cite{DiMP} and then estimate the bounding box of the target using the offline trained IoU module.
Despite the great improvements made by the above methods, they do not update all the sub-tasks (classification and box-estimation) online.
Instead, we propose a meta learnt context-aware model to update both the classifier and the bounding box estimator jointly, contributing to more robust and accurate tracking performance.

\vspace{2mm}
\noindent\textbf{Online adaption models.}
Since a target object undergoes various changes in a frame sequence, it is crucial to adjust the tracking model online to new target appearance. 
However, it is not straightforward to train a deep tracking model online since the online training samples are limited and the training needs high-computational loads.
To this end, effective adaption mechanisms~\cite{TADT,GradNet, CLNet, MAST} are explored.
GradNet~\cite{GradNet} updates the target template by exploiting discriminative information with the gradients through a gradient-guided network.
SiamAtt~\cite{SiamAtt} extracts contextual information between the target template and the search region for online adaption. 
In addition, meta-learning based models~\cite{Meta-tracker, MAML-Tracker, CLNet}, aiming to improve a learning algorithm over multiple learning episodes, are also used to accelerate the online training of existing trackers. 
Despite there are many works~\cite{MAML, MAML++, LEO, ProtoContext}, exploring meta-learning for few-shot classification tasks, meta-learning based tracking method are few in the visual tracking community.
Meta-Tracker~\cite{Meta-tracker} is a pioneering work, which devised an offline meta-learning based method to adjust the initial deep networks and applied it in MDNet~\cite{MDNET} and CREST~\cite{CREST}.
CLNet~\cite{CLNet} noticed the flaw in discrimination ability of Siamese trackers, for which a meta-learning based compact feature is proposed to capture the sequence-specific information at the first frame. But this method only uses the feature in the first frame, it is not sufficient to capture the information along the whole sequence, making it lack the capability of tracking robustly when target object appearance varies dramatically.
MAML-Tracker~\cite{MAML-Tracker} proposed a three-step approach based on MAML~\cite{MAML}, the typical gradient-based meta learning algorithm, to convert any modern detectors into instance level detectors for tracking. First the given detector is offline meta-trained to learn a parameter representation which can quickly adapt to tracking task. During online tracking, the model is first fine-tuned over the start frame, then optimized over subsequent frames of high confidence. Despite this method offers a general way to bridge the gap between tracking and detection, this method successively optimize ovet the parameter space of the classification and regression head during tracking, which is prone to over-fit even collapse on longer sequences. 
Different from these methods, we develop a meta-initialization scheme coupled with a context-aware model to jointly update the localization and box-estimation modules. During tracking, the joint updater is first fine-tuned on the start frame to capture target-specific information, then the joint meta-updater synchronously yields localization and regression affinity vectors to optimize the model on the representation space, such that these two sub-tasks can stably guide each other by exploiting the information from the other.

\begin{figure*}[ht]
\centering
\includegraphics[width=0.9\linewidth]{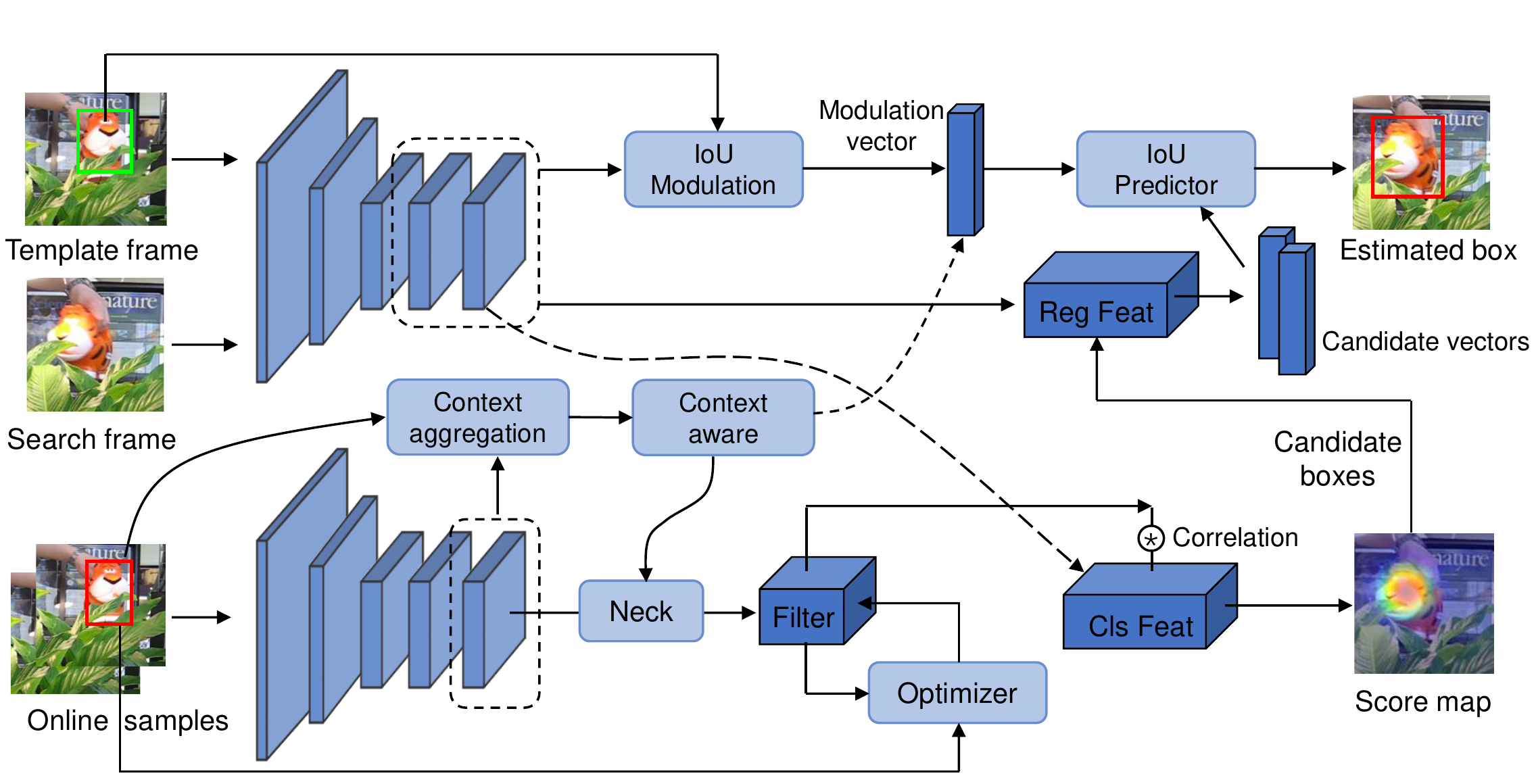}
\caption{\textbf{Overall framework.} It is composed of the base framework (backbone, ridge regression, and IoU modules) and the joint-update model including a context aggregation module and a context aware module.
The parameter of the context aggregation module is shared between the sub-tasks of classification and box-estimation, while the context-aware module specifies the ridge regression and IoU estimation tasks with different affinity vectors.}
\label{fig:framework}
\vspace{-2mm}
\end{figure*}

\section{Proposed Tracking Method}
Fig.~\ref{fig:framework} shows the overall framework of the proposed algorithm.
We construct a joint-update model under the DiMP framework.
The joint-update model shares the backbone with the DiMP framework and is comprised of a context-aggregation module and a context-aware module.
The context-aggregation module aggregates the features of multiple online samples and generates task-specific (localization and box estimation) vectors encoding target information.
The context-aware module then generates corresponding affinity vectors for adapting the localization and box estimation branches respectively.
In addition, we devise a target-specific meta-learning method to initialize the context-aware module based on the given initial target sample.

\subsection{Revisiting DiMP}
We choose DiMP~\cite{DiMP} comprised of a classification(ridge regression) branch and a box-estimation (IoU) branch as our baseline model.
The classification branch is implemented as a classifier with ridge regression method. Concretely, a filter kernel is initialized in the start frame with the given target annotation and is fine-tuned online with a dedicated optimizer every couple of frames.
The box-estimation branch, implemented as an overlap maximization method, is composed of an IoU modulation and an IoU predictor module that only trained offline.
During tracking, the classification branch first estimates the center position of the target and then the box-estimation branch predicts the target box based on several candidates drew near the predicted center position.
Adapting the IoU branch to the changing target object appearance is crucial for improving tracking accuracy, however, DiMP does not focus on this.
The IoU branch of DiMP is only initialized with the modulation vector using the initial target sample and not updated over time, which makes it less effective in handling new target appearances and inconsistent with the adaptive classification branch.
Besides, we find that there is more information in the sample memory that can be exploited to benefit the box-estimation branch and further boost the localization branch.
%
%
To this end, we first develop a context-aggregation module to encode the target-specific features of online samples and then devise a context-aware module to update both tasks via adaption vectors.
In the following, we will introduce these two modules in details.

\subsection{Context-aggregation module}
We observed that despite the target model(filter kernel) in the classification branch is well optimized during online tracking, the feature extractor is not stable to unseen target so that the classification feature of test frame is not optimal in online tracking.  
In the box estimation branch, the modulation vector is extracted only using the start frame target appearance and given annotations.
It is incapable of tracking target with various variation during online tracking, which will lead to inaccurate bounding box estimation.
We visualize some representative tracking results in Fig.(\ref{fig:vis_tracker}).
Candidate boxes of the box-estimation branch are generated according to precedent estimated target scales and predefined hyper-parameter. 
Without updating, the bounding box estimation error is accumulated during tracking a sequence, further may yield box with scale much larger than the target in some challenging cases, which introduce background clutter for the online optimized target model, contaminating the classification branch and then leading to severe drifting error.

\begin{figure}[t]
\centering
\includegraphics[width=1\linewidth]{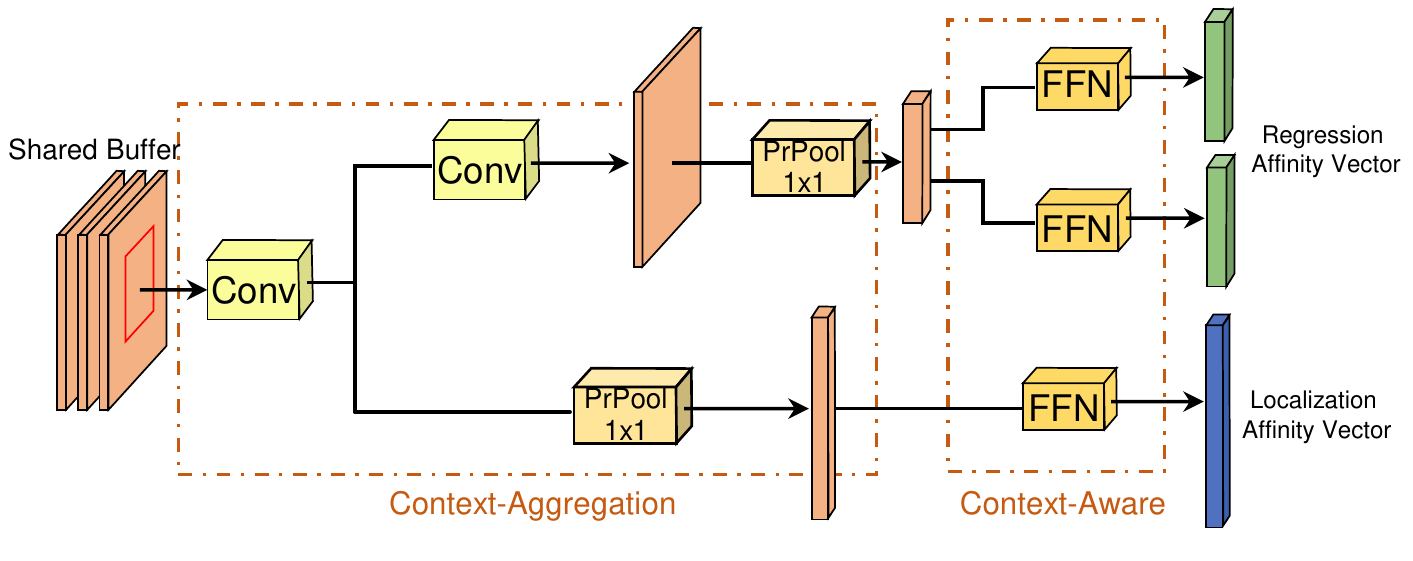}
\caption{Joint updater. Better viewed in color with zoom in.}
\label{fig:updater}
\vspace{-4mm}
\end{figure}

Considering the relationships between the tasks of classification and box estimation, we devise a context-aggregation module to learn the feature embeddings of both tasks on the compact representations.
The detailed structure of the context-aggregation module is shown in Fig.~\ref{fig:updater}.
The classification branch maintains a sample buffer in online tracking, can be denoted as $S_{online} = \{(x_i, b_i)\}_{i=1}^{n}$, which is composed of $n$ deep features $x_i$ paired with target box $b_i \in \mathbb R^4$
To make full use of these samples for both branches, we tried several SOTA attention-based method to learn a better representation for subsequent affinity vector generation, like Triple attention~\cite{Triplet_Att} and LamdaNetworks~\cite{lambdanetworks}. Despite these method show strong capability in other deep learning tasks like classification and detection, we find them achieve slight or without performance boost in our joint-updater. Based on this insight and for less computation burden, we adopt a simple convolutional layer here to extract compact representations from the online collected sample features (sample buffer).
Then we use a BatchNorm layer $u$ followed a ReLU activation to yield localization context feature from online samples, with which our context-aggregation module shares memory with the model predictor of the localization branch without any extra memory cost.
The localization feature is computed as:
\begin{eqnarray}
x_i^L = ReLU(u(x_i)), \,\, x_i \in S_{train} .
\end{eqnarray}
Here the the localization context feature is denoted as  $L_{online} = \{(x_i^L, b_i)\}_{i=1}^{n}$ . 

To effectively update the modulation vector, we use a $3\times 3$ convolution layer $g$ to extract regression context feature $R_{online} = \{(x_i^R, b_i)\}_{i=1}^n$ from $L_{online}$ as following:
\begin{eqnarray}
x_i^R=g(x_i^L),  \,\,\,\,  x_i^L \in L_{online}.
\end{eqnarray}
Ultimately, we use precise ROI pooling~\cite{prpool} to generate target-specific feature embeddings for both tasks, the target-specific representation for localization is computed as:
\begin{eqnarray}
l =\frac{1}{|L_{online}|} \sum_{i=1}^n PrPool(x_i^L, b_i),
\end{eqnarray}
where $PrPool$ denotes precise ROI pooling. The regression target-specific representation $r$ is derived from $R_{online}$ in the same way.

\subsection{Context-aware module}
\label{meta-learning}
In this part, we devise the context-aware module that transforms target-specific representation into task-specific affinity vector and a meta initialization scheme is further proposed to ensure stable and fast updating for both branches.
Given the localization-specific and regression-specific representations $l_j$ and $r_j$ of the target in sequence $j$, we pass them to fully connected layer followed with a $sigmoid$ activation function, which transforms $l_j$ and $r_j$ into localization affinity vector $A_j^L$, and regression affinity vector $A_j^{R_i}$, respectively. 
The elements in these affinity vectors indicate how important each channel is for the localization and regression sub-tasks.
Specifically, we use the localization affinity vector $A_j^L$ to adapt the localization feature as following:
\begin{eqnarray}
s = (x\odot A_j^L)*f,
\end{eqnarray}
where $\odot$ denotes the element-wise multiplication in the channel dimension , $x$ is the extracted localization feature of test frame, and $f$ is the target model(filter weight) maintained by the classification branch, $*$ is correlation operator.
Similarly, the two regression affinity vectors $A_j^{R_1}$ and $A_j^{R_2}$ are respectively used to adapt the modulation vectors $M_j^0$ and $M_j^1$ extracted by the IoU modulation module from the start frame with given annotation.
With updated modulation vector, the IoU scores of candidate bounding boxes are computed as:
\begin{eqnarray}
IoU(B)=h(z(x_{j_i}, B), {M_j^0\odot A_j^{R_0}, M_j^1\odot A_j^{R_1}}),
\end{eqnarray}
where $z$ denotes the feature extractor of the test frame, $h$ denotes the IoU predictor in the bounding box regression branch.

\subsection{Meta-initialization}
Considering the different characteristics of different object categories, we devise a meta-learning scheme to adapt the context-aware module to a specific target object. 

MAML~\cite{MAML} is the typical meta-learning method and widely used in visual tracking problems in previous work. 
When MAML-like meta-learning algorithm coupled with tracking method, there is a drawback that the offline meta-trained representation which can quickly adapt to new tasks is prone to over-fitting. In other words, the offline learned model parameter space is polluted if successive online updating is directly applied in the parameter space. 
That's why MAML-Tracker~\cite{MAML-Tracker},optimizing the parameter of classification and regression head during online tracking with MAML, shows poor performance on large-scale dataset like LaSOT~\cite{LaSOT}.

From this perspective, we adapt the original MAML algorithm and use the new scheme to train and adapt the context-aware part of the joint meta-updater in start frame. When tracking the subsequent frames, our method optimizes the model in high-dimensional representation space rather than the parameter space, avoid the overfitting problem when gradient-based meta-learning coupled with tracking model.

Specifically, we divide the parameter of the meta-updater into two parts.
The context-aggregation module with parameter $\phi$ is acted as sequence-shared parameter, which remains invariant during online tracking, while the parameter of the context-aware module $\theta$ is online updated. 
For offline training, we sample frames $D_j=\{x_j^i, b_j^i\}$ from $j$-$th$ video sequences $V_j$ then divide them into $D_{j}^{s}$ (support set) and $D_j^{t}$ (target set). 
Alg.~\ref{Alg:meta_train} presents the detailed optimization process.
%
In the inner loop optimization, we adopt multi-step gradient-descent with respect to context-aware parameter $\theta$, minimizing the loss $L$ on $D_j^{s}$ , here $p$ denotes localization or state estimation module combined with the meta-updater.
The process can be formulated as:
\begin{equation}
     \theta_k = \theta_{k-1}-\alpha\odot \frac{\partial }{\partial \theta_{k-1}}M(\theta_{k-1}, \phi),
     \label{innerloop_update}
\end{equation}
and $ M(\theta_{k-1}, \phi ;D_j^{s})$ is computed as:
\begin{equation}
 M(\theta_{k-1}, \phi ;D_j^{s}) = \frac{1}{|D_j^{s}|}\sum_{(x_j^i,b_j^i)\in{D_j^{s}}} L(p(x_j^i;\theta_{k-1},\phi),b_j^i).
\end{equation}
The offline training loss $L$ defined as the weighted sum of IoU loss of box-estimation branch and hinge regression loss in the classification branch. The loss in classification branch for ridge regression is define as:

\begin{equation}
L_{cls} = \frac{1}{|D_j|} \sum_{(x_j^i, b_j^i) \in D_j)} \Vert \ell (s, z(b_j^i)) \Vert^2,
\end{equation}
here $f$ is the filter weight, z denotes a transformation from bounding box $b_j^i$ to a gaussian label for ridge regression. $l$ is a hinge loss between the predicted score map $s$ and gaussian label $z$:
\begin{equation}
\ell(s, z)=\left\{\begin{array}{ll}
s-z, & z>T \\
\max (0, s), & z \leq T,
\end{array}\right.
\end{equation}
$T$ is a pre-defined threshold to exclude distractors in the background. The loss $L_{IoU}$ in regression branch is computed as between the estimated overlap and the ground truth. The total loss is the sum of these two branches with a balance hyper-parameter $\lambda$:
\begin{equation}
    L = \lambda L_{cls} + L_{IoU}.
\end{equation}
In the outer-loop optimization, the loss definition is same as in the inner-loop, to stable the training, the test losses on $K$ step optimized $\{\theta_{k}\}_{k=1}^K$ are fused in a weighted sum manner.
The process is formulated as:
\begin{eqnarray}
T(\theta, \phi; D_j^{t}) = \frac{1}{|D_j^{t}|}\sum_{k=0}^K \sum_{(x_j^i, b_j^i)}v_k\,L(p(x_j^i;\theta_{k-1},\phi), b_j^i).
\label{eq:test_loss}
\end{eqnarray}
Finally, the weighted test loss $T$ is backpropagated to find a global optimal representation $\{\theta, \phi\}^*$ that can quickly adapt to new target appearance as:
\begin{eqnarray}
\{\theta, \phi\}^* = \arg\min_{\{{\theta, \phi\}}}\frac{1}{N}\sum_{j=0}^{N}T(\theta, \phi; D_j^{t}).
\label{eq:outer-loop}
\end{eqnarray}
Here $N$ denote the total sampled video sequences in a training batch, and $v_k$ denote the loss weight of the $k$-${th}$ step, which grows with step number to pay more attention to later optimization steps.
During online adaption, only the parameter of context aware module $\theta$ is updated by $N$ step optimization using \eqref{innerloop_update} over frames, while the parameter of context aggregation module $\phi$ is fixed as an initialization. 
In this setting, our model can adapt the meta-updater to an unseen target appearance efficiently.

\begin{algorithm}[t]
\small
    \caption{Meta initialization training}
    \label{Alg:meta_train}
    \LinesNumbered
    \KwIn{Video sequences set ${V_j}$ of frames paired with annotated bounding box and a pretrained model of the first-stage\;}
    \KwOut{Optimized representation $\{\theta, \phi\}^*$ capable of fast adapting to new target appearance\;}
    { Draw image batches from different videos to form a training set $D_j$, then split the training set into support set $D_j^s$ and test set $D_j^t$\;}
    \While{not done}{
    \For {$k=1, ..., N$}{
    Evaluate $M(\theta_{k-1}, \phi, D_j^s)$ with respect to $|D_j^s|$ using \eqref{innerloop_update}\;
    Update the context aware parameter $\theta$ with gradient descent:$\theta_k =\theta_{k-1} - \alpha \nabla_{\theta_{k-1}}{M}$\;
    }
    Compute $T(\theta, \phi, D_j^t)$ on test set for each optimized $\theta_k$ using \eqref{eq:test_loss}, then perform back-propagation to find the optimal representation using \eqref{eq:outer-loop}.}
\end{algorithm}

\section{Experiment}
In this section, we evaluate the proposed algorithm on 4 diverse datasets and compare it with 10 representative methods.
In addition, we conduct an ablation study to show the contribution of each component of the proposed algorithm.
In the followings, we first introduce the implementations, and then present the state-of-the-art comparison and the ablation study.

\subsection{Implementation details}
\textbf{Network Architecture. } We use the vanilla ResNet50~\cite{RES50} as the feature extractor, the search region and template share the same backbone. The last layer of ResNet50 is removed, the feature of layer3 and layer4 is extracted then passed into the IoU-modulation module to produce two modulation vectors of length 256. The classification branch maintains a filter kernel with spatial size $3*3$, the channel dimension is 512. The context-aggregation module is implemented with a convolution followed by a batch normalization and a ReLU activation. For the localization affinity vector, the output of the aggregation module is directly pooled by a $1*1$ PrPool into a vector then passed into the feed-forward network.
For the regression affinity vector, the output of aggregation module is downsampled by another convolution, then pooled by a $1*1$ PrPool and fed into the feed-forward network to yield the regression affinity vector.

\vspace{2mm}
\noindent\textbf{Offline training. }
For offline training, we first train the overall framework including the joint update model and the DiMP model in an end-to-end manner. 
We use the ADAM optimization method with a learning rate of 1e-4 and weight decay for training. 
The loss is defined as the sum of the localization and box estimation branches. 
In the second stage, we train the model obtained in the first stage to learn meta-initialization parameters.
The training loss is the same as that of the DiMP but using a meta-learning strategy. 
For the meta training, the inner-loop iteration number is set as 5.
In the outer loop, we assigned the weights $\{v_i\}_{i=0}^5 = \{0.1, 0.2, 0.4, 0.6, 0.8, 1\}$ to iteration steps, giving more attention to later steps. 
To stable the training, we use learnable per-layer-per-step inner loop learning rate initialized as 0.01. 
We train 20 epochs using the ADAM optimizer with learning rates of 5e-5 on the context-aware module and 1e-5 on the context-aggregation module. 
In these two training stages, we use TrackignNet, LaSOT, Got10k and Coco as training dataset and batch size is set as 40.
The total training process takes 13 hours on a server with 4 V100 Tesla GPUs and a Xeon 5118s CPU.

\vspace{2mm}
\noindent\textbf{Online tracking. }For online tracking, the classification branch maintains a sample buffer of maximum size 50 by discarding the oldest samples. In the start frame, data augmentation including blur, rotation, translation and dropout are performed to generate 15 augmented samples to initialize the filter weight of in the classification branch and the modulation vectors in the box-estimation branch. Then we run 5 iterations on these augmented samples with their given annotation to adapt the context-aware part of our joint updater to the specified target appearance. When tracking subsequent frames, the weight of the meta-joint updater is fixed, the sample buffer is updated with high confidence new frames. The affinity vector of both branches is updated synchronously with the sample buffer. The average tracking speed is 40FPS on a machine with a 1080ti GPU and an i7-9700k CPU, only with a slight speed drop compared to the baseline DiMP50 of 45FPS.

\subsection{Comparison with SOTA}

We compare our proposed method with the state-of-the-art methods on four challenging datasets including VOT2018, OTB100, NFS, TrackingNet and LaSOT.
We show qualitative evaluation of our proposed method and baseline DiMP50 on several challenging sequences in Fig.~\ref{fig:vis_tracker}. A tracking video demo can be found in the supplementary materials.
%


\begin{table}[htbp]
\vspace{-4mm}
  \centering
  \caption{Results on the VOT2018 dataset}
    \begin{tabular}{l|ccc}
    \toprule
          & EAO  $\uparrow$ & Rob  $\downarrow$ & ACC  $\uparrow$ \\
    \midrule
    DaSiamRPN~\cite{DaSiamRPN} & 0.383  & 0.276  & 0.586  \\
    UPDT~\cite{UPDT} & 0.378  & 0.184  & 0.536  \\
    UpdateNet~\cite{UpdateNet} & 0.393  & 0.276  & 0.587  \\
    TADT~\cite{TADT} & 0.224  & 0.487  & 0.502  \\
    SiamRPN++~\cite{SiamRPN++} & 0.414  & 0.234  & 0.600  \\
    ATOM~\cite{ATOM} & 0.401 & 0.204 & 0.590  \\
    CAGCD~\cite{CAGCD} & 0.449 & 0.173 & 0.615 \\
    SiamFC++~\cite{siamfc++} & 0.426 & 0.183 & 0.587 \\
    DiMP18~\cite{DiMP} & 0.402 & 0.182 & 0.594 \\
    DiMP50~\cite{DiMP} & 0.440  & \textcolor[rgb]{ .169,  .231,  .843}{0.153} & 0.597 \\
    SiamBAN~\cite{SiamBAN} & \textcolor[rgb]{ .169,  .231,  .843}{0.452} & 0.178 & 0.597 \\
    MAML-Tracker~\cite{MAML-Tracker} & \textcolor[rgb]{ .169,  .231,  .843}{0.452} & 0.159 & 0.604 \\
    SiamKPN~\cite{SiamKPN} & 0.440  & 0.192 & \textcolor[rgb]{ .169,  .231,  .843}{0.606} \\
    \midrule
    \textbf{CAJMU(ours)} & \textcolor[rgb]{ 1,  0,  0}{0.514} & \textcolor[rgb]{ 1,  0,  0}{0.108} & \textcolor[rgb]{ 1,  0,  0}{0.609} \\
    \bottomrule
    \end{tabular}%
  \label{tab:vot_result}%
  \vspace{-2mm}
\end{table}%

\vspace{2mm}
\noindent\textbf{VOT2018~\cite{VOT2018}}: 
This dataset is composed of 60 sequences, the target in sequences in annotated with rotated bounding box.
In this benchmark, trackers are assumed to be incapable of re-detection after the target has been lost. 
Once the overlap of the estimated box and corresponding ground truth box is 0, the tracker will be reinitialized using ground-truth label of five frames after the failure.
Three metrics is used to report the tracking performance: robustness, accuracy and  EAO. 
The accuracy is a metric to evaluate the average overlap between the estimated box and ground truth box when tracking successfully. The robustness measures how many times the tracker loses the target.
EAO is the primary metric based on robustness and accuracy metric, which is an estimator of short-term sequences with the same visual properties as the given dataset.
We compare our proposed method with 13 representative deep trackers on this dataset, Table~\ref{tab:vot_result} shows the results of all the methods over these three metrics. Otherwise, we plot the EAO scores of the totally 13 trackers in Fig.\ref{fig:eao_subplot}.
It shows that our method achieves the best performance with an EAO score of 0.514 and a relative gain of 16.8\% against DiMP50.

\begin{figure}[b]
\centering
\includegraphics[width=1\linewidth]{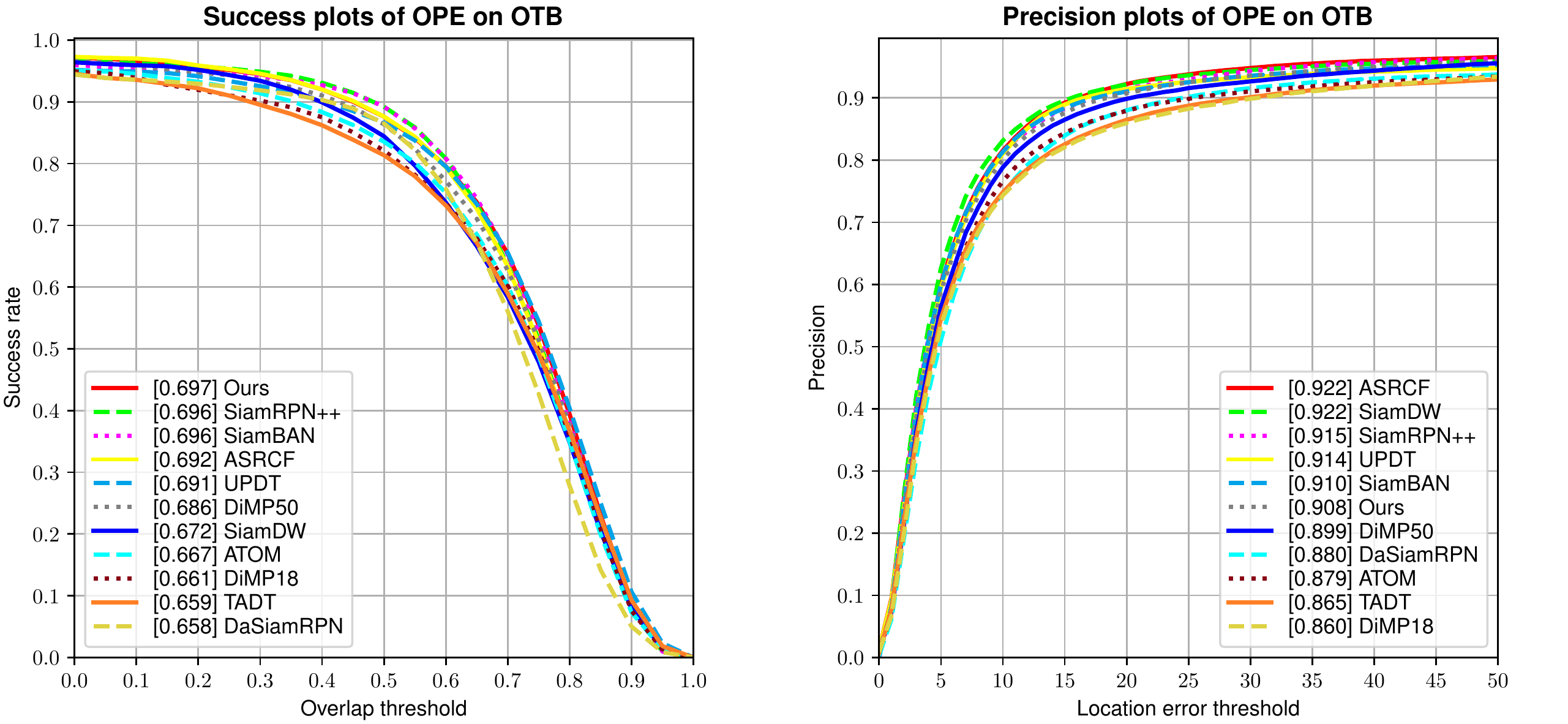}
\caption{\textbf{Success and Precision plot on OTB100 benchmark.} }
\label{fig:otb_subplot}
\end{figure}
\vspace{2mm}
\noindent\textbf{OTB100~\cite{OTB2015}:} This is one of the most widely used dataset in visual tracking community, which contains 100 sequences with various targets. Precision and success are the commonly used evaluation metrics in this dataset. The precision score is computed by comparing the difference between the estimated state and the ground-truth bounding box. The success metric represents the box Intersection of Union(IoU) of the predicted result and ground truth. Trackers are ranked by the Area Under Curve(AUC) scores. We compare our proposed method with 10 representative deep trackers, Fig.\ref{fig:otb_subplot} shows the detailed success and precision plot of the evaluation result. SiamRPN++~\cite{SiamRPN++} achieves the highest precision score among the compared methods. The proposed method achieves an AUC score of 0.697, improving DiMP50 with a relative gain of 1.9\%.

\begin{figure}[t]
\centering
\includegraphics[width=1\linewidth]{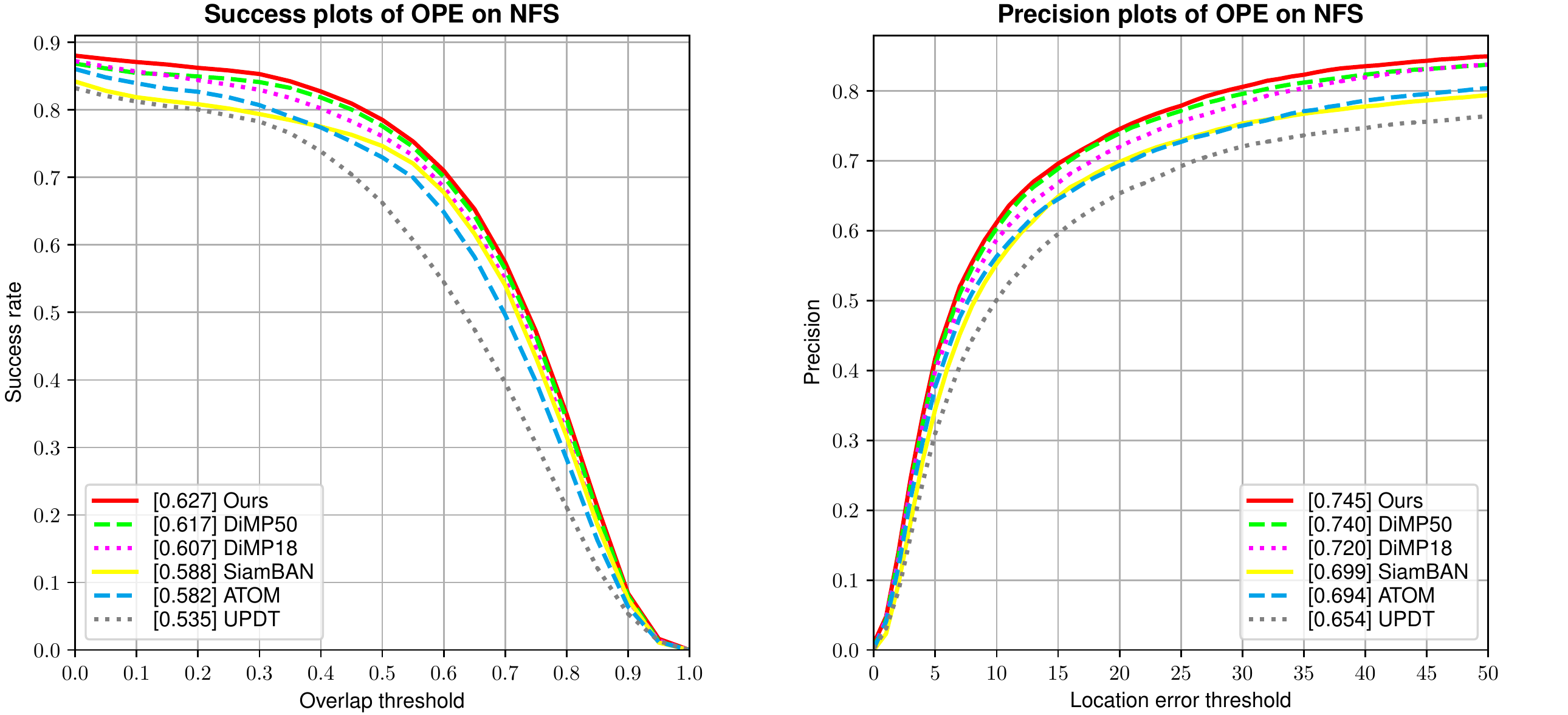}
\caption{\textbf{Success and Precision plot on NFS benchmark.} }
\label{fig:nfs_plot}
\vspace{-4mm}
\end{figure}

\vspace{2mm}
\noindent\textbf{NFS~\cite{NFS}:} The evaluation is performed on the 30FPS version of this benchmark. It is composed of 100 sequences captured with higher frame rate(240FPS) cameras from real world, the 30FPS version of this dataset is temporally sampled every 8th frame from the 240FPS version, which leading to excluding the effect of motion blur in lower frame tracking benchmark like OTB100. The evaluation metrics are the same as those in the OTB100 dataset, the Area Under Curve(AUC) of success plots are used to rank trackers. We compare our proposed method with 5 deep trackers, Fig.\ref{fig:nfs_plot} shows the success and precision plot. The DiMP50 baseline model obtains an AUC score of 0.620. Our approach achieves slightly better results with an AUC score of 0.627 and a precision score of 0.747.

\vspace{2mm}
\noindent\textbf{TrackingNet~\cite{TrackingNet}}: This dataset is the first large-scale benchmark for visual object tracking in the wild. It is composed of 511 sequences from YouTube, the annotations defined as the smallest upright bounding box to avoid ambiguity. Three metrics are used to provide the performance analysis: precision, normalized precision and success. Since the precision metric is sensitive to the resolution of images and the scale of bounding boxes. Precision is normalized over the size of the ground truth bounding box, which ensure the consistency of the metrics across different scales of targets to track. We compare our proposed method with 6 representative deep trackers on this benchmark, including DiMP50~\cite{DiMP}, SiamRPN++~\cite{SiamRPN++}, DaSiamRPN~\cite{DaSiamRPN}, ATOM~\cite{ATOM}, UPDT~\cite{UPDT} and SiamFC~\cite{SiamFC}, Table \ref{tab:trackingnet} shows detailed evaluation result. Our proposed method achieves slight improvement on precision and Success metric, with an AUC score of 74.2.

\begin{table}[htbp]
  \centering
  \caption{Result on the TrackingNet dataset}
\resizebox{1\linewidth}{!}{
    \begin{tabular}{lccccccc}
    \toprule
          & SiamFC & UPDT  & ATOM  & DaSiamRPN & SiamRPN++ & DiMP50 & \textbf{CAJMU(ours)} \\
    \midrule
    Precision (\%) & 53.3  & 55.7  & 64.8  & 59.1  & 69.4  & 68.7  & \textcolor[rgb]{ 1,  0,  0}{68.9 } \\
    Norm. Prec. (\%) & 66.6  & 70.2  & 77.1  & 73.3  & \textcolor[rgb]{ .169,  .231,  .843}{80.0 } & \textcolor[rgb]{ 1,  0,  0}{80.1 } & \textcolor[rgb]{ 1,  0,  0}{80.1 } \\
    Success (\%) & 57.1  & 61.1  & 70.3  & 63.8  & \textcolor[rgb]{ .169,  .231,  .843}{73.3 } & 74.0  & \textcolor[rgb]{ 1,  0,  0}{74.2 } \\
    \bottomrule
    \end{tabular}%
  \label{tab:trackingnet}
 }
\end{table}%

\begin{figure*}[t]
\centering
\includegraphics[width=0.98\linewidth]{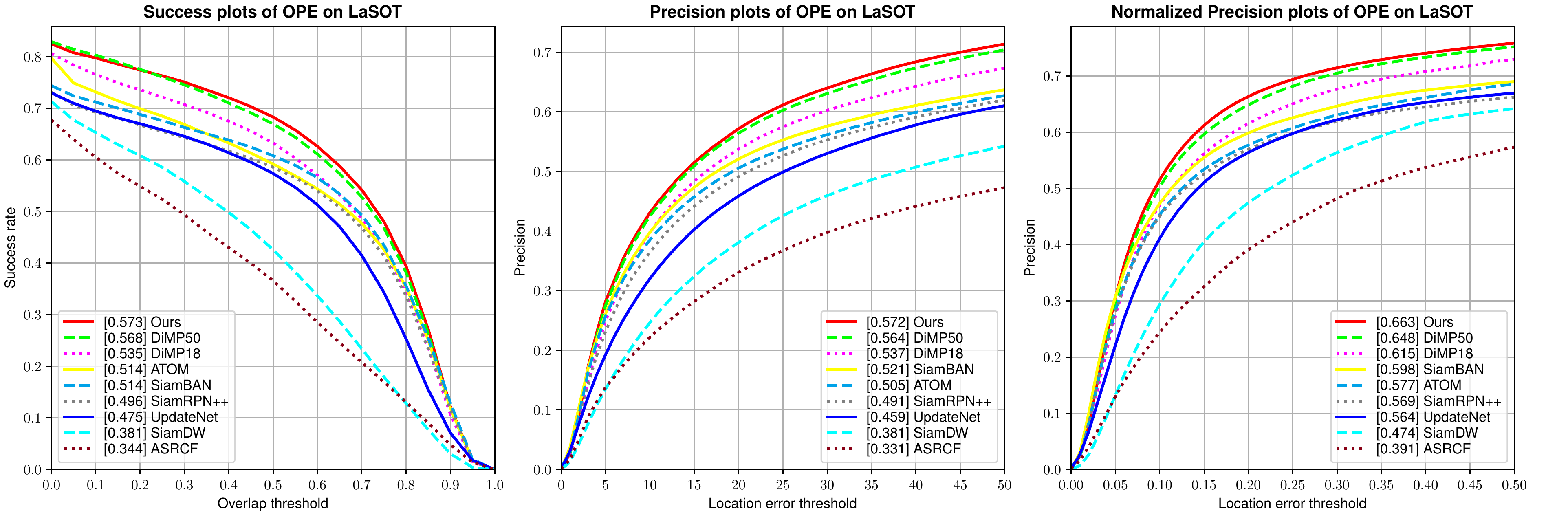}
\caption{\textbf{Results on the LaSOT benchmark.} }
\label{fig:lasot_subplot}
\end{figure*}

\vspace{2mm}
\noindent\textbf{LaSOT~\cite{LaSOT}}: This benchmark is the largest densely annotated one in the tracking community. To further evaluate our proposed method on challenging sequences, we also conduct experiment on the LaSOT dataset consisting 280 longer sequences compared to VOT2018 and OTB100, which totally contain 690K frames and 70 object categories. Following popular benchmark protocols, this benchmark also perform One-Pass Evaluation(OPE), and use the success and precision metric. Besides success and precision, LaSOT also introduces the normalized precision metric to alleviate the instability of precision metric. We compare our proposed method with 9 deep trackers on this benchmark, the curve of success, precision and normalized precision are depict in Fig.\ref{fig:lasot_subplot}. Our approach achieves the best performance over three metrics with an AUC score of 0.573, demonstrating its strong capability to tracking objects with various variation in large-scale benchmark.

The extensive experimental results on these 5 diverse datasets demonstrate that our approach performs favorably against the state-of-the-art methods.

\subsection{Ablation study}
In this section, we explore the contribution of each component in the proposed joint meta-updater over the OTB100 and VOT2018 benchmarks. 
Table \ref{ablation_study} presents the results of four variants of the proposed model. First we only update localization(cls) or regression(reg) branch. 
When we only update the localization(cls) branch with the localization affinity vector, our tracker still can achieve an EAO score of 0.473 on VOT2018 benchmark, showing that the compact representation in localization branch is capable of optimizing the localization search feature. When only the regression affinity vector is applied, our model also can boost the tracking performance on both benchmarks. This experiment result also shows that the update of the regression branch has more contribution on OTB100 dataset, but has less effect on VOT2018 compared to the update in localization branch, since target appearance in VOT2018 benchmark varies more strenuous than the OTB100 benchmark. Then we jointly update both branches without meta initialization. Benefiting from the joint update mechanism, we get an EAO score of 0.506 on VOT2018 and an AUC score of 0.686 on OTB100, achieving significant performance gain compared to update single branch, which demonstrates that updating in these two branches are supplementary to each other. 

At last, we combine our joint updater with the meta-initialization scheme, achieving performance boost on both benchmarks. The precision metric on OTB100 and accuracy metric on VOT2018 are both lifted, showing our meta-initialization scheme is capable of stabilizing and further boosting the performance.

\begin{table}[tbp]
  \centering
  \caption{\textbf{Ablation study on OTB100 and VOT2018}}
 \resizebox{0.95\linewidth}{!}{   \begin{tabular}{lll|ccccc}
    \toprule
          &       &       & \multicolumn{2}{c}{OTB100 } & \multicolumn{3}{c}{VOT2018 } \\
    cls   & reg   & adapt & AUC $\uparrow$ & Pre.  $\uparrow$ & EAO  $\uparrow$ & Rob.  $\downarrow$ & Acc.  $\uparrow$ \\
    \midrule
    \checkmark &       &       & 0.686  & 0.891  & 0.473  & 0.117  & 0.598  \\
          & \checkmark &       & 0.689  & 0.895  & 0.455  & 0.150  & \textcolor[rgb]{ .169,  .231,  .843}{0.602}  \\
    \checkmark & \checkmark &       & \textcolor[rgb]{ .169,  .231,  .843}{0.694}  & \textcolor[rgb]{ .169,  .231,  .843}{0.904}  & \textcolor[rgb]{ .169,  .231,  .843}{0.506}  & \textcolor[rgb]{ 1,  0,  0}{0.103}  & 0.601  \\
    \checkmark & \checkmark & \checkmark & \textcolor[rgb]{ 1,  0,  0}{0.697}  & \textcolor[rgb]{ 1,  0,  0}{0.908}  & \textcolor[rgb]{ 1,  0,  0}{0.514}  & \textcolor[rgb]{ .169,  .231,  .843}{0.108}  &\textcolor[rgb]{ 1,  0,  0}{0.609}  \\
    \bottomrule
    \end{tabular} }
  \label{ablation_study}%
\vspace{-2mm}
\end{table}%

\section{Conclusions}
In this paper, we propose a joint meta-updater to adapt the deep tracking model to target objects fast and stably. Our key insight lies in that the information of the classification branch and the box-estimation branch can complement each other. As such, we construct a context aggregation and a context aware module for jointly generating task-specific affinity vectors for both branches. To stabilize the performance of our joint updater, we develop a dedicated meta initialization scheme to ensure a fast adaption of our joint updater by exploiting target-specific information. In general, we provide a novel way to update the tracking model on the representation space rather than directly on the parameter space, which can make full use of the information along the whole sequence without over-fitting problem. The extensive experiments on five public benchmarks show that the proposed algorithm performs favorably against the state-of-the-art method.

\clearpage
\bibliographystyle{ACM-Reference-Format}
\bibliography{cajmu}


\begin{thebibliography}{51}


\ifx \showCODEN    \undefined \def \showCODEN     #1{\unskip}     \fi
\ifx \showDOI      \undefined \def \showDOI       #1{#1}\fi
\ifx \showISBNx    \undefined \def \showISBNx     #1{\unskip}     \fi
\ifx \showISBNxiii \undefined \def \showISBNxiii  #1{\unskip}     \fi
\ifx \showISSN     \undefined \def \showISSN      #1{\unskip}     \fi
\ifx \showLCCN     \undefined \def \showLCCN      #1{\unskip}     \fi
\ifx \shownote     \undefined \def \shownote      #1{#1}          \fi
\ifx \showarticletitle \undefined \def \showarticletitle #1{#1}   \fi
\ifx \showURL      \undefined \def \showURL       {\relax}        \fi
\providecommand\bibfield[2]{#2}
\providecommand\bibinfo[2]{#2}
\providecommand\natexlab[1]{#1}
\providecommand\showeprint[2][]{arXiv:#2}

\bibitem[\protect\citeauthoryear{Antoniou, Edwards, and Storkey}{Antoniou
  et~al\mbox{.}}{2018}]%
        {MAML++}
\bibfield{author}{\bibinfo{person}{Antreas Antoniou}, \bibinfo{person}{Harrison
  Edwards}, {and} \bibinfo{person}{Amos Storkey}.}
  \bibinfo{year}{2018}\natexlab{}.
\newblock \showarticletitle{How to train your maml}.
\newblock \bibinfo{journal}{\emph{arXiv preprint arXiv:1810.09502}}
  (\bibinfo{year}{2018}).
\newblock


\bibitem[\protect\citeauthoryear{Bello}{Bello}{2021}]%
        {lambdanetworks}
\bibfield{author}{\bibinfo{person}{Irwan Bello}.}
  \bibinfo{year}{2021}\natexlab{}.
\newblock \showarticletitle{Lambdanetworks: Modeling long-range interactions
  without attention}.
\newblock \bibinfo{journal}{\emph{arXiv preprint arXiv:2102.08602}}
  (\bibinfo{year}{2021}).
\newblock


\bibitem[\protect\citeauthoryear{Bertinetto, Valmadre, Henriques, Vedaldi, and
  Torr}{Bertinetto et~al\mbox{.}}{2016}]%
        {SiamFC}
\bibfield{author}{\bibinfo{person}{Luca Bertinetto}, \bibinfo{person}{Jack
  Valmadre}, \bibinfo{person}{Joao~F Henriques}, \bibinfo{person}{Andrea
  Vedaldi}, {and} \bibinfo{person}{Philip~HS Torr}.}
  \bibinfo{year}{2016}\natexlab{}.
\newblock \showarticletitle{Fully-convolutional siamese networks for object
  tracking}. In \bibinfo{booktitle}{\emph{European conference on computer
  vision}}. Springer, \bibinfo{pages}{850--865}.
\newblock


\bibitem[\protect\citeauthoryear{Bhat, Danelljan, Van~Gool, and Timofte}{Bhat
  et~al\mbox{.}}{2019}]%
        {DiMP}
\bibfield{author}{\bibinfo{person}{Goutam Bhat}, \bibinfo{person}{Martin
  Danelljan}, \bibinfo{person}{Luc Van~Gool}, {and} \bibinfo{person}{Radu
  Timofte}.} \bibinfo{year}{2019}\natexlab{}.
\newblock \showarticletitle{Learning Discriminative Model Prediction for
  Tracking}. In \bibinfo{booktitle}{\emph{IEEE International Conference on
  Computer Vision}}.
\newblock


\bibitem[\protect\citeauthoryear{Bhat, Johnander, Danelljan, Khan, and
  Felsberg}{Bhat et~al\mbox{.}}{2018}]%
        {UPDT}
\bibfield{author}{\bibinfo{person}{Goutam Bhat}, \bibinfo{person}{Joakim
  Johnander}, \bibinfo{person}{Martin Danelljan},
  \bibinfo{person}{Fahad~Shahbaz Khan}, {and} \bibinfo{person}{Michael
  Felsberg}.} \bibinfo{year}{2018}\natexlab{}.
\newblock \showarticletitle{Unveiling the Power of Deep Tracking}. In
  \bibinfo{booktitle}{\emph{European Conference on Computer Vision}}.
\newblock


\bibitem[\protect\citeauthoryear{Chen, Yan, Zhu, Wang, Yang, and Lu}{Chen
  et~al\mbox{.}}{2021}]%
        {TransT}
\bibfield{author}{\bibinfo{person}{Xin Chen}, \bibinfo{person}{Bin Yan},
  \bibinfo{person}{Jiawen Zhu}, \bibinfo{person}{Dong Wang},
  \bibinfo{person}{Xiaoyun Yang}, {and} \bibinfo{person}{Huchuan Lu}.}
  \bibinfo{year}{2021}\natexlab{}.
\newblock \showarticletitle{Transformer Tracking}.
\newblock \bibinfo{journal}{\emph{arXiv preprint arXiv:2103.15436}}
  (\bibinfo{year}{2021}).
\newblock


\bibitem[\protect\citeauthoryear{Chen, Zhong, Li, Zhang, and Ji}{Chen
  et~al\mbox{.}}{2020}]%
        {SiamBAN}
\bibfield{author}{\bibinfo{person}{Zedu Chen}, \bibinfo{person}{Bineng Zhong},
  \bibinfo{person}{Guorong Li}, \bibinfo{person}{Shengping Zhang}, {and}
  \bibinfo{person}{Rongrong Ji}.} \bibinfo{year}{2020}\natexlab{}.
\newblock \showarticletitle{Siamese Box Adaptive Network for Visual Tracking}.
  In \bibinfo{booktitle}{\emph{Proceedings of the IEEE/CVF Conference on
  Computer Vision and Pattern Recognition}}. \bibinfo{pages}{6668--6677}.
\newblock


\bibitem[\protect\citeauthoryear{Choi, Kwon, and Lee}{Choi
  et~al\mbox{.}}{2020}]%
        {TACT}
\bibfield{author}{\bibinfo{person}{Janghoon Choi}, \bibinfo{person}{Junseok
  Kwon}, {and} \bibinfo{person}{Kyoung~Mu Lee}.}
  \bibinfo{year}{2020}\natexlab{}.
\newblock \showarticletitle{Visual Tracking by TridentAlign and Context
  Embedding}. In \bibinfo{booktitle}{\emph{Proceedings of the Asian Conference
  on Computer Vision}}.
\newblock


\bibitem[\protect\citeauthoryear{Danelljan, Bhat, Khan, and Felsberg}{Danelljan
  et~al\mbox{.}}{2017}]%
        {ECO}
\bibfield{author}{\bibinfo{person}{Martin Danelljan}, \bibinfo{person}{Goutam
  Bhat}, \bibinfo{person}{F~Shahbaz Khan}, {and} \bibinfo{person}{Michael
  Felsberg}.} \bibinfo{year}{2017}\natexlab{}.
\newblock \showarticletitle{ECO: Efficient convolution operators for tracking}.
  In \bibinfo{booktitle}{\emph{IEEE Conference on Computer Vision and Pattern
  Recognition}}.
\newblock


\bibitem[\protect\citeauthoryear{Danelljan, Bhat, Khan, and Felsberg}{Danelljan
  et~al\mbox{.}}{2019}]%
        {ATOM}
\bibfield{author}{\bibinfo{person}{Martin Danelljan}, \bibinfo{person}{Goutam
  Bhat}, \bibinfo{person}{Fahad~Shahbaz Khan}, {and} \bibinfo{person}{Michael
  Felsberg}.} \bibinfo{year}{2019}\natexlab{}.
\newblock \showarticletitle{Atom: Accurate tracking by overlap maximization}.
  In \bibinfo{booktitle}{\emph{IEEE Conference on Computer Vision and Pattern
  Recognition}}.
\newblock


\bibitem[\protect\citeauthoryear{Danelljan, Gool, and Timofte}{Danelljan
  et~al\mbox{.}}{2020}]%
        {PrDiMP}
\bibfield{author}{\bibinfo{person}{Martin Danelljan}, \bibinfo{person}{Luc~Van
  Gool}, {and} \bibinfo{person}{Radu Timofte}.}
  \bibinfo{year}{2020}\natexlab{}.
\newblock \showarticletitle{Probabilistic Regression for Visual Tracking}. In
  \bibinfo{booktitle}{\emph{IEEE Conference on Computer Vision and Pattern
  Recognition}}.
\newblock


\bibitem[\protect\citeauthoryear{Danelljan, Robinson, Khan, and
  Felsberg}{Danelljan et~al\mbox{.}}{2016}]%
        {CCOT}
\bibfield{author}{\bibinfo{person}{Martin Danelljan}, \bibinfo{person}{Andreas
  Robinson}, \bibinfo{person}{Fahad~Shahbaz Khan}, {and}
  \bibinfo{person}{Michael Felsberg}.} \bibinfo{year}{2016}\natexlab{}.
\newblock \showarticletitle{Beyond correlation filters: Learning continuous
  convolution operators for visual tracking}. In
  \bibinfo{booktitle}{\emph{European Conference on Computer Vision}}.
\newblock


\bibitem[\protect\citeauthoryear{Dong, Shen, Shao, and Porikli}{Dong
  et~al\mbox{.}}{2020}]%
        {CLNet}
\bibfield{author}{\bibinfo{person}{Xingping Dong}, \bibinfo{person}{Jianbing
  Shen}, \bibinfo{person}{Ling Shao}, {and} \bibinfo{person}{Fatih Porikli}.}
  \bibinfo{year}{2020}\natexlab{}.
\newblock \showarticletitle{CLNet: A compact latent network for fast adjusting
  Siamese trackers}. In \bibinfo{booktitle}{\emph{Proceedings of the 16th
  European Conference on Computer Vision}}. Springer,
  \bibinfo{pages}{378--395}.
\newblock


\bibitem[\protect\citeauthoryear{Du, Liu, Zhao, and Tang}{Du
  et~al\mbox{.}}{2020}]%
        {CAGCD}
\bibfield{author}{\bibinfo{person}{Fei Du}, \bibinfo{person}{Peng Liu},
  \bibinfo{person}{Wei Zhao}, {and} \bibinfo{person}{Xianglong Tang}.}
  \bibinfo{year}{2020}\natexlab{}.
\newblock \showarticletitle{Correlation-Guided Attention for Corner Detection
  Based Visual Tracking}. In \bibinfo{booktitle}{\emph{Proceedings of the
  IEEE/CVF Conference on Computer Vision and Pattern Recognition}}.
  \bibinfo{pages}{6836--6845}.
\newblock


\bibitem[\protect\citeauthoryear{Fan, Lin, Yang, Chu, Deng, Yu, Bai, Xu, Liao,
  and Ling}{Fan et~al\mbox{.}}{2019}]%
        {LaSOT}
\bibfield{author}{\bibinfo{person}{Heng Fan}, \bibinfo{person}{Liting Lin},
  \bibinfo{person}{Fan Yang}, \bibinfo{person}{Peng Chu}, \bibinfo{person}{Ge
  Deng}, \bibinfo{person}{Sijia Yu}, \bibinfo{person}{Hexin Bai},
  \bibinfo{person}{Yong Xu}, \bibinfo{person}{Chunyuan Liao}, {and}
  \bibinfo{person}{Haibin Ling}.} \bibinfo{year}{2019}\natexlab{}.
\newblock \showarticletitle{Lasot: A high-quality benchmark for large-scale
  single object tracking}. In \bibinfo{booktitle}{\emph{IEEE Conference on
  Computer Vision and Pattern Recognition}}.
\newblock


\bibitem[\protect\citeauthoryear{Fan and Ling}{Fan and Ling}{2019}]%
        {C-RPN}
\bibfield{author}{\bibinfo{person}{Heng Fan} {and} \bibinfo{person}{Haibin
  Ling}.} \bibinfo{year}{2019}\natexlab{}.
\newblock \showarticletitle{Siamese cascaded region proposal networks for
  real-time visual tracking}. In \bibinfo{booktitle}{\emph{IEEE Conference on
  Computer Vision and Pattern Recognition}}.
\newblock


\bibitem[\protect\citeauthoryear{Finn, Abbeel, and Levine}{Finn
  et~al\mbox{.}}{2017}]%
        {MAML}
\bibfield{author}{\bibinfo{person}{Chelsea Finn}, \bibinfo{person}{Pieter
  Abbeel}, {and} \bibinfo{person}{Sergey Levine}.}
  \bibinfo{year}{2017}\natexlab{}.
\newblock \showarticletitle{Model-agnostic meta-learning for fast adaptation of
  deep networks}. In \bibinfo{booktitle}{\emph{International Conference on
  Machine Learning}}. PMLR, \bibinfo{pages}{1126--1135}.
\newblock


\bibitem[\protect\citeauthoryear{Galoogahi, Fagg, Huang, Ramanan, and
  Lucey}{Galoogahi et~al\mbox{.}}{2017}]%
        {NFS}
\bibfield{author}{\bibinfo{person}{Hamed~Kiani Galoogahi},
  \bibinfo{person}{Ashton Fagg}, \bibinfo{person}{Chen Huang},
  \bibinfo{person}{Deva Ramanan}, {and} \bibinfo{person}{Simon Lucey}.}
  \bibinfo{year}{2017}\natexlab{}.
\newblock \showarticletitle{Need for Speed: A Benchmark for Higher Frame Rate
  Object Tracking}.
\newblock \bibinfo{journal}{\emph{arXiv preprint arXiv:1703.05884}}
  (\bibinfo{year}{2017}).
\newblock


\bibitem[\protect\citeauthoryear{Grigorescu, Trasnea, Cocias, and
  Macesanu}{Grigorescu et~al\mbox{.}}{2020}]%
        {ad_survey}
\bibfield{author}{\bibinfo{person}{Sorin Grigorescu}, \bibinfo{person}{Bogdan
  Trasnea}, \bibinfo{person}{Tiberiu Cocias}, {and} \bibinfo{person}{Gigel
  Macesanu}.} \bibinfo{year}{2020}\natexlab{}.
\newblock \showarticletitle{A survey of deep learning techniques for autonomous
  driving}.
\newblock \bibinfo{journal}{\emph{Journal of Field Robotics}}
  \bibinfo{volume}{37}, \bibinfo{number}{3} (\bibinfo{year}{2020}),
  \bibinfo{pages}{362--386}.
\newblock


\bibitem[\protect\citeauthoryear{Guo, Shao, Cui, Wang, Zhang, and Shen}{Guo
  et~al\mbox{.}}{2020}]%
        {graph_track}
\bibfield{author}{\bibinfo{person}{Dongyan Guo}, \bibinfo{person}{Yanyan Shao},
  \bibinfo{person}{Ying Cui}, \bibinfo{person}{Zhenhua Wang},
  \bibinfo{person}{Liyan Zhang}, {and} \bibinfo{person}{Chunhua Shen}.}
  \bibinfo{year}{2020}\natexlab{}.
\newblock \showarticletitle{Graph Attention Tracking}.
\newblock \bibinfo{journal}{\emph{arXiv preprint arXiv:2011.11204}}
  (\bibinfo{year}{2020}).
\newblock


\bibitem[\protect\citeauthoryear{He, Zhang, Ren, and Sun}{He
  et~al\mbox{.}}{2016}]%
        {RES50}
\bibfield{author}{\bibinfo{person}{Kaiming He}, \bibinfo{person}{Xiangyu
  Zhang}, \bibinfo{person}{Shaoqing Ren}, {and} \bibinfo{person}{Jian Sun}.}
  \bibinfo{year}{2016}\natexlab{}.
\newblock \showarticletitle{Deep residual learning for image recognition}. In
  \bibinfo{booktitle}{\emph{IEEE Conference on Computer Vision and Pattern
  Recognition}}.
\newblock


\bibitem[\protect\citeauthoryear{Jiang, Luo, Mao, Xiao, and Jiang}{Jiang
  et~al\mbox{.}}{2018}]%
        {prpool}
\bibfield{author}{\bibinfo{person}{Borui Jiang}, \bibinfo{person}{Ruixuan Luo},
  \bibinfo{person}{Jiayuan Mao}, \bibinfo{person}{Tete Xiao}, {and}
  \bibinfo{person}{Yuning Jiang}.} \bibinfo{year}{2018}\natexlab{}.
\newblock \showarticletitle{Acquisition of localization confidence for accurate
  object detection}. In \bibinfo{booktitle}{\emph{Proceedings of the European
  Conference on Computer Vision (ECCV)}}. \bibinfo{pages}{784--799}.
\newblock


\bibitem[\protect\citeauthoryear{Koohzadi and Charkari}{Koohzadi and
  Charkari}{2017}]%
        {ac_survey}
\bibfield{author}{\bibinfo{person}{Maryam Koohzadi} {and}
  \bibinfo{person}{Nasrollah~Moghadam Charkari}.}
  \bibinfo{year}{2017}\natexlab{}.
\newblock \showarticletitle{Survey on deep learning methods in human action
  recognition}.
\newblock \bibinfo{journal}{\emph{IET Computer Vision}} \bibinfo{volume}{11},
  \bibinfo{number}{8} (\bibinfo{year}{2017}), \bibinfo{pages}{623--632}.
\newblock


\bibitem[\protect\citeauthoryear{Kristan, Leonardis, Matas, Felsberg,
  Pflugfelder, ˇCehovin~Zajc, Vojir, Bhat, Lukezic, Eldesokey,
  et~al\mbox{.}}{Kristan et~al\mbox{.}}{2018}]%
        {VOT2018}
\bibfield{author}{\bibinfo{person}{Matej Kristan}, \bibinfo{person}{Ales
  Leonardis}, \bibinfo{person}{Jiri Matas}, \bibinfo{person}{Michael Felsberg},
  \bibinfo{person}{Roman Pflugfelder}, \bibinfo{person}{Luka ˇCehovin~Zajc},
  \bibinfo{person}{Tomas Vojir}, \bibinfo{person}{Goutam Bhat},
  \bibinfo{person}{Alan Lukezic}, \bibinfo{person}{Abdelrahman Eldesokey},
  {et~al\mbox{.}}} \bibinfo{year}{2018}\natexlab{}.
\newblock \showarticletitle{The sixth visual object tracking vot2018 challenge
  results}. In \bibinfo{booktitle}{\emph{European Conference on Computer Vision
  Workshops}}.
\newblock


\bibitem[\protect\citeauthoryear{Lai, Lu, and Xie}{Lai et~al\mbox{.}}{2020}]%
        {MAST}
\bibfield{author}{\bibinfo{person}{Zihang Lai}, \bibinfo{person}{Erika Lu},
  {and} \bibinfo{person}{Weidi Xie}.} \bibinfo{year}{2020}\natexlab{}.
\newblock \showarticletitle{MAST: A memory-augmented self-supervised tracker}.
  In \bibinfo{booktitle}{\emph{Proceedings of the IEEE/CVF Conference on
  Computer Vision and Pattern Recognition}}. \bibinfo{pages}{6479--6488}.
\newblock


\bibitem[\protect\citeauthoryear{Li, Wu, Wang, Zhang, Xing, and Yan}{Li
  et~al\mbox{.}}{2018a}]%
        {SiamRPN++}
\bibfield{author}{\bibinfo{person}{Bo Li}, \bibinfo{person}{Wei Wu},
  \bibinfo{person}{Qiang Wang}, \bibinfo{person}{Fangyi Zhang},
  \bibinfo{person}{Junliang Xing}, {and} \bibinfo{person}{Junjie Yan}.}
  \bibinfo{year}{2018}\natexlab{a}.
\newblock \showarticletitle{SiamRPN++: Evolution of Siamese Visual Tracking
  with Very Deep Networks}. In \bibinfo{booktitle}{\emph{IEEE Conference on
  Computer Vision and Pattern Recognition}}.
\newblock


\bibitem[\protect\citeauthoryear{Li, Yan, Wu, Zhu, and Hu}{Li
  et~al\mbox{.}}{2018b}]%
        {SiamRPN}
\bibfield{author}{\bibinfo{person}{Bo Li}, \bibinfo{person}{Junjie Yan},
  \bibinfo{person}{Wei Wu}, \bibinfo{person}{Zheng Zhu}, {and}
  \bibinfo{person}{Xiaolin Hu}.} \bibinfo{year}{2018}\natexlab{b}.
\newblock \showarticletitle{High Performance Visual Tracking With Siamese
  Region Proposal Network}. In \bibinfo{booktitle}{\emph{IEEE Conference on
  Computer Vision and Pattern Recognition}}.
\newblock


\bibitem[\protect\citeauthoryear{Li, Chen, Ouyang, Wang, Yang, and Lu}{Li
  et~al\mbox{.}}{2019a}]%
        {GradNet}
\bibfield{author}{\bibinfo{person}{Peixia Li}, \bibinfo{person}{Boyu Chen},
  \bibinfo{person}{Wanli Ouyang}, \bibinfo{person}{Dong Wang},
  \bibinfo{person}{Xiaoyun Yang}, {and} \bibinfo{person}{Huchuan Lu}.}
  \bibinfo{year}{2019}\natexlab{a}.
\newblock \showarticletitle{GradNet: Gradient-Guided Network for Visual Object
  Tracking}. In \bibinfo{booktitle}{\emph{IEEE International Conference on
  Computer Vision}}.
\newblock


\bibitem[\protect\citeauthoryear{Li, Qin, Zhang, and Zheng}{Li
  et~al\mbox{.}}{2020}]%
        {SiamKPN}
\bibfield{author}{\bibinfo{person}{Qiang Li}, \bibinfo{person}{Zekui Qin},
  \bibinfo{person}{Wenbo Zhang}, {and} \bibinfo{person}{Wen Zheng}.}
  \bibinfo{year}{2020}\natexlab{}.
\newblock \showarticletitle{Siamese Keypoint Prediction Network for Visual
  Object Tracking}.
\newblock \bibinfo{journal}{\emph{arXiv preprint arXiv:2006.04078}}
  (\bibinfo{year}{2020}).
\newblock


\bibitem[\protect\citeauthoryear{Li, Ma, Wu, He, and Yang}{Li
  et~al\mbox{.}}{2019b}]%
        {TADT}
\bibfield{author}{\bibinfo{person}{Xin Li}, \bibinfo{person}{Chao Ma},
  \bibinfo{person}{Baoyuan Wu}, \bibinfo{person}{Zhenyu He}, {and}
  \bibinfo{person}{Ming-Hsuan Yang}.} \bibinfo{year}{2019}\natexlab{b}.
\newblock \showarticletitle{Target-Aware Deep Tracking}. In
  \bibinfo{booktitle}{\emph{IEEE Conference on Computer Vision and Pattern
  Recognition}}.
\newblock


\bibitem[\protect\citeauthoryear{{Liu}, {Li}, {He}, {Fan}, {Yuan}, and
  {Wang}}{{Liu} et~al\mbox{.}}{2020}]%
        {DMSTIT}
\bibfield{author}{\bibinfo{person}{Q. {Liu}}, \bibinfo{person}{X. {Li}},
  \bibinfo{person}{Z. {He}}, \bibinfo{person}{N. {Fan}}, \bibinfo{person}{D.
  {Yuan}}, {and} \bibinfo{person}{H. {Wang}}.} \bibinfo{year}{2020}\natexlab{}.
\newblock \showarticletitle{Learning Deep Multi-Level Similarity for Thermal
  Infrared Object Tracking}.
\newblock \bibinfo{journal}{\emph{IEEE Transactions on Multimedia}}
  (\bibinfo{year}{2020}), \bibinfo{pages}{1--1}.
\newblock
\urldef\tempurl%
\url{https://doi.org/10.1109/TMM.2020.3008028}
\showDOI{\tempurl}


\bibitem[\protect\citeauthoryear{Ma, Wang, Zhang, Lu, and Yin}{Ma
  et~al\mbox{.}}{2020}]%
        {RPT}
\bibfield{author}{\bibinfo{person}{Ziang Ma}, \bibinfo{person}{Linyuan Wang},
  \bibinfo{person}{Haitao Zhang}, \bibinfo{person}{Wei Lu}, {and}
  \bibinfo{person}{Jun Yin}.} \bibinfo{year}{2020}\natexlab{}.
\newblock \showarticletitle{RPT: Learning Point Set Representation for Siamese
  Visual Tracking}.
\newblock \bibinfo{journal}{\emph{arXiv preprint arXiv:2008.03467}}
  (\bibinfo{year}{2020}).
\newblock


\bibitem[\protect\citeauthoryear{Misra, Nalamada, Arasanipalai, and Hou}{Misra
  et~al\mbox{.}}{2021}]%
        {Triplet_Att}
\bibfield{author}{\bibinfo{person}{Diganta Misra}, \bibinfo{person}{Trikay
  Nalamada}, \bibinfo{person}{Ajay~Uppili Arasanipalai}, {and}
  \bibinfo{person}{Qibin Hou}.} \bibinfo{year}{2021}\natexlab{}.
\newblock \showarticletitle{Rotate to attend: Convolutional triplet attention
  module}. In \bibinfo{booktitle}{\emph{Proceedings of the IEEE/CVF Winter
  Conference on Applications of Computer Vision}}. \bibinfo{pages}{3139--3148}.
\newblock


\bibitem[\protect\citeauthoryear{Muller, Bibi, Giancola, Alsubaihi, and
  Ghanem}{Muller et~al\mbox{.}}{2018}]%
        {TrackingNet}
\bibfield{author}{\bibinfo{person}{Matthias Muller}, \bibinfo{person}{Adel
  Bibi}, \bibinfo{person}{Silvio Giancola}, \bibinfo{person}{Salman Alsubaihi},
  {and} \bibinfo{person}{Bernard Ghanem}.} \bibinfo{year}{2018}\natexlab{}.
\newblock \showarticletitle{Trackingnet: A large-scale dataset and benchmark
  for object tracking in the wild}. In \bibinfo{booktitle}{\emph{European
  Conference on Computer Vision}}.
\newblock


\bibitem[\protect\citeauthoryear{Nam and Han}{Nam and Han}{2016}]%
        {MDNET}
\bibfield{author}{\bibinfo{person}{Hyeonseob Nam} {and}
  \bibinfo{person}{Bohyung Han}.} \bibinfo{year}{2016}\natexlab{}.
\newblock \showarticletitle{Learning multi-domain convolutional neural networks
  for visual tracking}. In \bibinfo{booktitle}{\emph{IEEE Conference on
  Computer Vision and Pattern Recognition}}.
\newblock


\bibitem[\protect\citeauthoryear{Park and Berg}{Park and Berg}{2018}]%
        {Meta-tracker}
\bibfield{author}{\bibinfo{person}{Eunbyung Park} {and}
  \bibinfo{person}{Alexander~C Berg}.} \bibinfo{year}{2018}\natexlab{}.
\newblock \showarticletitle{Meta-tracker: Fast and robust online adaptation for
  visual object trackers}. In \bibinfo{booktitle}{\emph{Proceedings of the
  European Conference on Computer Vision (ECCV)}}. \bibinfo{pages}{569--585}.
\newblock


\bibitem[\protect\citeauthoryear{Rusu, Rao, Sygnowski, Vinyals, Pascanu,
  Osindero, and Hadsell}{Rusu et~al\mbox{.}}{2018}]%
        {LEO}
\bibfield{author}{\bibinfo{person}{Andrei~A Rusu}, \bibinfo{person}{Dushyant
  Rao}, \bibinfo{person}{Jakub Sygnowski}, \bibinfo{person}{Oriol Vinyals},
  \bibinfo{person}{Razvan Pascanu}, \bibinfo{person}{Simon Osindero}, {and}
  \bibinfo{person}{Raia Hadsell}.} \bibinfo{year}{2018}\natexlab{}.
\newblock \showarticletitle{Meta-learning with latent embedding optimization}.
\newblock \bibinfo{journal}{\emph{arXiv preprint arXiv:1807.05960}}
  (\bibinfo{year}{2018}).
\newblock


\bibitem[\protect\citeauthoryear{Shih}{Shih}{2017}]%
        {va_survey}
\bibfield{author}{\bibinfo{person}{Huang-Chia Shih}.}
  \bibinfo{year}{2017}\natexlab{}.
\newblock \showarticletitle{A survey of content-aware video analysis for
  sports}.
\newblock \bibinfo{journal}{\emph{IEEE Transactions on Circuits and Systems for
  Video Technology}} \bibinfo{volume}{28}, \bibinfo{number}{5}
  (\bibinfo{year}{2017}), \bibinfo{pages}{1212--1231}.
\newblock


\bibitem[\protect\citeauthoryear{Song, Ma, Gong, Zhang, Lau, and Yang}{Song
  et~al\mbox{.}}{2017}]%
        {CREST}
\bibfield{author}{\bibinfo{person}{Yibing Song}, \bibinfo{person}{Chao Ma},
  \bibinfo{person}{Lijun Gong}, \bibinfo{person}{Jiawei Zhang},
  \bibinfo{person}{Rynson~WH Lau}, {and} \bibinfo{person}{Ming-Hsuan Yang}.}
  \bibinfo{year}{2017}\natexlab{}.
\newblock \showarticletitle{Crest: Convolutional residual learning for visual
  tracking}. In \bibinfo{booktitle}{\emph{IEEE International Conference on
  Computer Vision}}. \bibinfo{pages}{2574--2583}.
\newblock


\bibitem[\protect\citeauthoryear{Vogelbaum, Dangovski, Jing, and
  Solja{\v{c}}i{\'c}}{Vogelbaum et~al\mbox{.}}{2020}]%
        {ProtoContext}
\bibfield{author}{\bibinfo{person}{Evan Vogelbaum}, \bibinfo{person}{Rumen
  Dangovski}, \bibinfo{person}{Li Jing}, {and} \bibinfo{person}{Marin
  Solja{\v{c}}i{\'c}}.} \bibinfo{year}{2020}\natexlab{}.
\newblock \showarticletitle{Contextualizing Enhances Gradient Based Meta
  Learning}.
\newblock \bibinfo{journal}{\emph{arXiv preprint arXiv:2007.10143}}
  (\bibinfo{year}{2020}).
\newblock


\bibitem[\protect\citeauthoryear{Voigtlaender, Luiten, Torr, and
  Leibe}{Voigtlaender et~al\mbox{.}}{2020}]%
        {SiamR-CNN}
\bibfield{author}{\bibinfo{person}{Paul Voigtlaender},
  \bibinfo{person}{Jonathon Luiten}, \bibinfo{person}{Philip~H.S. Torr}, {and}
  \bibinfo{person}{Bastian Leibe}.} \bibinfo{year}{2020}\natexlab{}.
\newblock \showarticletitle{Siam R-CNN: Visual Tracking by Re-Detection}. In
  \bibinfo{booktitle}{\emph{IEEE Conference on Computer Vision and Pattern
  Recognition}}.
\newblock


\bibitem[\protect\citeauthoryear{Wang, Luo, Sun, Xiong, and Zeng}{Wang
  et~al\mbox{.}}{2020}]%
        {MAML-Tracker}
\bibfield{author}{\bibinfo{person}{Guangting Wang}, \bibinfo{person}{Chong
  Luo}, \bibinfo{person}{Xiaoyan Sun}, \bibinfo{person}{Zhiwei Xiong}, {and}
  \bibinfo{person}{Wenjun Zeng}.} \bibinfo{year}{2020}\natexlab{}.
\newblock \showarticletitle{Tracking by instance detection: A meta-learning
  approach}. In \bibinfo{booktitle}{\emph{Proceedings of the IEEE/CVF
  Conference on Computer Vision and Pattern Recognition}}.
  \bibinfo{pages}{6288--6297}.
\newblock


\bibitem[\protect\citeauthoryear{Wang, Gao, Xing, Zhang, and Hu}{Wang
  et~al\mbox{.}}{2017}]%
        {dcfnet}
\bibfield{author}{\bibinfo{person}{Qiang Wang}, \bibinfo{person}{Jin Gao},
  \bibinfo{person}{Junliang Xing}, \bibinfo{person}{Mengdan Zhang}, {and}
  \bibinfo{person}{Weiming Hu}.} \bibinfo{year}{2017}\natexlab{}.
\newblock \showarticletitle{Dcfnet: Discriminant correlation filters network
  for visual tracking}.
\newblock \bibinfo{journal}{\emph{arXiv preprint arXiv:1704.04057}}
  (\bibinfo{year}{2017}).
\newblock


\bibitem[\protect\citeauthoryear{Wu, Lim, and Yang}{Wu et~al\mbox{.}}{2015}]%
        {OTB2015}
\bibfield{author}{\bibinfo{person}{Yi Wu}, \bibinfo{person}{Jongwoo Lim}, {and}
  \bibinfo{person}{Ming-Hsuan Yang}.} \bibinfo{year}{2015}\natexlab{}.
\newblock \showarticletitle{Object tracking benchmark}.
\newblock \bibinfo{journal}{\emph{IEEE Transactions on Pattern Analysis and
  Machine Intelligence}} \bibinfo{volume}{37}, \bibinfo{number}{9}
  (\bibinfo{year}{2015}), \bibinfo{pages}{1834--1848}.
\newblock


\bibitem[\protect\citeauthoryear{Xu, Wang, Li, Yuan, and Yu}{Xu
  et~al\mbox{.}}{2020}]%
        {siamfc++}
\bibfield{author}{\bibinfo{person}{Yinda Xu}, \bibinfo{person}{Zeyu Wang},
  \bibinfo{person}{Zuoxin Li}, \bibinfo{person}{Ye Yuan}, {and}
  \bibinfo{person}{Gang Yu}.} \bibinfo{year}{2020}\natexlab{}.
\newblock \showarticletitle{Siamfc++: Towards robust and accurate visual
  tracking with target estimation guidelines}. In
  \bibinfo{booktitle}{\emph{Proceedings of the AAAI Conference on Artificial
  Intelligence}}, Vol.~\bibinfo{volume}{34}. \bibinfo{pages}{12549--12556}.
\newblock


\bibitem[\protect\citeauthoryear{Yang, Xu, Hu, Chai, and Chan}{Yang
  et~al\mbox{.}}{2020}]%
        {ROAM}
\bibfield{author}{\bibinfo{person}{Tianyu Yang}, \bibinfo{person}{Pengfei Xu},
  \bibinfo{person}{Runbo Hu}, \bibinfo{person}{Hua Chai}, {and}
  \bibinfo{person}{Antoni~B Chan}.} \bibinfo{year}{2020}\natexlab{}.
\newblock \showarticletitle{ROAM: Recurrently optimizing tracking model}. In
  \bibinfo{booktitle}{\emph{Proceedings of the IEEE/CVF Conference on Computer
  Vision and Pattern Recognition}}. \bibinfo{pages}{6718--6727}.
\newblock


\bibitem[\protect\citeauthoryear{Yu, Xiong, Huang, and Scott}{Yu
  et~al\mbox{.}}{2020}]%
        {SiamAtt}
\bibfield{author}{\bibinfo{person}{Yuechen Yu}, \bibinfo{person}{Yilei Xiong},
  \bibinfo{person}{Weilin Huang}, {and} \bibinfo{person}{Matthew~R. Scott}.}
  \bibinfo{year}{2020}\natexlab{}.
\newblock \showarticletitle{Deformable Siamese Attention Networks for Visual
  Object Tracking}. In \bibinfo{booktitle}{\emph{IEEE Conference on Computer
  Vision and Pattern Recognition}}.
\newblock


\bibitem[\protect\citeauthoryear{Zhang, Gonzalez-Garcia, Weijer, Danelljan, and
  Khan}{Zhang et~al\mbox{.}}{2019}]%
        {UpdateNet}
\bibfield{author}{\bibinfo{person}{Lichao Zhang}, \bibinfo{person}{Abel
  Gonzalez-Garcia}, \bibinfo{person}{Joost van~de Weijer},
  \bibinfo{person}{Martin Danelljan}, {and} \bibinfo{person}{Fahad~Shahbaz
  Khan}.} \bibinfo{year}{2019}\natexlab{}.
\newblock \showarticletitle{Learning the Model Update for Siamese Trackers}. In
  \bibinfo{booktitle}{\emph{The IEEE International Conference on Computer
  Vision (ICCV)}}.
\newblock


\bibitem[\protect\citeauthoryear{Zhang and Peng}{Zhang and Peng}{2019}]%
        {SiamDW}
\bibfield{author}{\bibinfo{person}{Zhipeng Zhang} {and} \bibinfo{person}{Houwen
  Peng}.} \bibinfo{year}{2019}\natexlab{}.
\newblock \showarticletitle{Deeper and Wider Siamese Networks for Real-Time
  Visual Tracking}. In \bibinfo{booktitle}{\emph{The IEEE Conference on
  Computer Vision and Pattern Recognition (CVPR)}}.
\newblock


\bibitem[\protect\citeauthoryear{Zhao, Zhang, Ma, Tang, Wang, and Wang}{Zhao
  et~al\mbox{.}}{2020}]%
        {SGT}
\bibfield{author}{\bibinfo{person}{Fei Zhao}, \bibinfo{person}{Ting Zhang},
  \bibinfo{person}{Chao Ma}, \bibinfo{person}{Ming Tang},
  \bibinfo{person}{Jinqiao Wang}, {and} \bibinfo{person}{Xiaobo Wang}.}
  \bibinfo{year}{2020}\natexlab{}.
\newblock \showarticletitle{Siamese attentive graph tracking}. In
  \bibinfo{booktitle}{\emph{Proceedings of the 28th ACM International
  Conference on Multimedia}}. \bibinfo{pages}{1542--1550}.
\newblock


\bibitem[\protect\citeauthoryear{Zhu, Wang, Li, Wu, Yan, and Hu}{Zhu
  et~al\mbox{.}}{2018}]%
        {DaSiamRPN}
\bibfield{author}{\bibinfo{person}{Zheng Zhu}, \bibinfo{person}{Qiang Wang},
  \bibinfo{person}{Bo Li}, \bibinfo{person}{Wei Wu}, \bibinfo{person}{Junjie
  Yan}, {and} \bibinfo{person}{Weiming Hu}.} \bibinfo{year}{2018}\natexlab{}.
\newblock \showarticletitle{Distractor-aware Siamese Networks for Visual Object
  Tracking}. In \bibinfo{booktitle}{\emph{European Conference on Computer
  Vision}}.
\newblock


\end{thebibliography}
\end{document}